\documentclass{article} 
\usepackage{iclr2025_conference,times}

\usepackage{amsmath,amsfonts,bm}

\def\eqref#1{equation~\ref{#1}}

\def\1{\bm{1}}

\DeclareMathAlphabet{\mathsfit}{\encodingdefault}{\sfdefault}{m}{sl}
\SetMathAlphabet{\mathsfit}{bold}{\encodingdefault}{\sfdefault}{bx}{n}

\usepackage{placeins}

\usepackage{hyperref}
\usepackage{url}
\usepackage{graphicx}
\usepackage{subfigure}
\usepackage{caption}
\usepackage[ruled,vlined]{algorithm2e}
\usepackage{booktabs}
\usepackage{colortbl}
\colorlet{tableheadcolor}{cyan!25} 
\newcommand{\headcol}{\rowcolor{tableheadcolor}} %
\colorlet{tablerowcolor}{cyan!10} 

\title{Predicting 3D structure by latent posterior sampling}

\author{Azmi Haider \\
Department of Computer Science\\
University of Haifa\\
Haifa, Israel \\
\texttt{ahaide03@campus.haifa.ac.il} \\
\And
Dan Rosenbaum \\
Department of Computational Science \\
University of Haifa\\
Haifa, Israel \\
\texttt{danro@cs.haifa.ac.il} \\
}

\iclrfinalcopy 
\begin{document}

\maketitle


\begin{abstract}
The remarkable achievements of both generative models of 2D images and neural field representations for 3D scenes present a compelling opportunity to integrate the strengths of both approaches.
In this work, we propose a methodology that combines a NeRF-based representation of 3D scenes with probabilistic modeling and reasoning using diffusion models.
We view 3D reconstruction as a perception problem with inherent uncertainty that can thereby benefit from probabilistic inference methods.  
The core idea is to represent the 3D scene as a stochastic latent variable for which we can learn a prior and use it to perform posterior inference given a set of observations. 
We formulate posterior sampling using the score-based inference method of diffusion models in conjunction with a likelihood term computed from a reconstruction model that includes volumetric rendering. 
We train the model using a two-stage process: first we train the reconstruction model while auto-decoding the latent representations for a dataset of 3D scenes, and then we train the prior over the latents using a diffusion model.
By using the model to generate samples from the posterior we demonstrate that various 3D reconstruction tasks can be performed, differing by the type of observation used as inputs. 
We showcase reconstruction from single-view, multi-view, noisy images, sparse pixels, and sparse depth data. 
These observations vary in the amount of information they provide for the scene and we show that our method can model the varying levels of inherent uncertainty associated with each task.
Our experiments illustrate that this approach yields a comprehensive method capable of accurately predicting 3D structure from diverse types of observations.
\end{abstract}


\section{Introduction}
\label{sec:introduction}
    


\begin{figure}[ht]
\centering
\scriptsize{\hspace{1.4cm} observation \; sample 1 \; sample 2 \; sample 3 \; uncertainty 
\hspace{1.1cm} observation \; sample 1 \; sample 2 \; sample 3 \; uncertainty}  
\begin{tabular}{cc}
  \begin{tabular}{c}
    half \\ image \\[3.5cm] \end{tabular} &
    \includegraphics[width=0.87\textwidth]{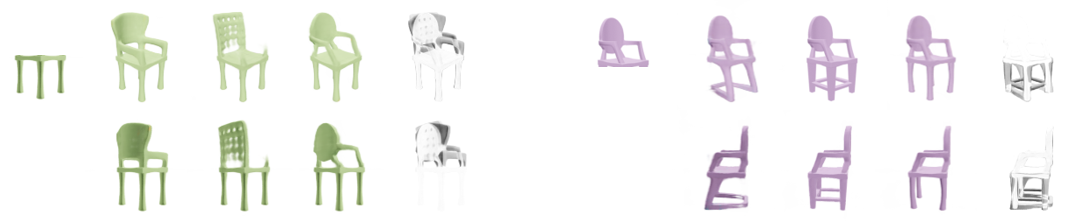} \vspace{-1.8cm}  \\
  
  \begin{tabular}{c}
    sparse \\ pixels \\[3.5cm] \end{tabular} &  
  \includegraphics[width=0.87\textwidth]{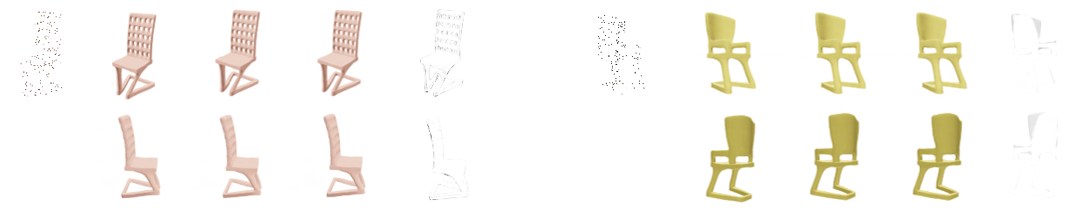} \vspace{-1.8cm} \\
  \begin{tabular}{c}
    saprse \\ depth \\[3.5cm] \end{tabular} &
  \includegraphics[width=0.87\textwidth]{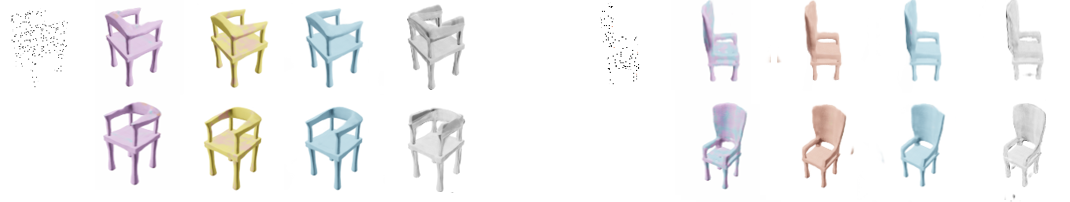} \vspace{-1.8cm}
\end{tabular}
  \caption{Examples of various 3D prediction tasks performed by generating posterior samples with our method. For each task we show the observation, three samples of the scene and an uncertainty map computed from the variance of 10 samples. Two different views are shown in subsequent rows. \textbf{Top:} reconstruction from half of an image. The variance is high in the hidden half of the scene. \textbf{Middle:} reconstruction from only a few pixels (5\% of a single image). \textbf{Bottom:} reconstruction from a few depth values (5\% of a full depth image from a single direction). Samples and uncertainty map suggest sparse depth is enough to reconstruct the 3D shape and uncertainty remains only about color. }
\label{fig:partialguidance}
\end{figure}

3D prediction using neural networks ~\citep{mildenhall2020nerf, sitzmann2019srns, park2019deepsdflearningcontinuoussigned} has garnered significant attention, tackling two main challenges: \textbf{3D reconstruction} (predicting 3D representations from limited observations) and \textbf{3D generation} (sampling new 3D scenes using generative models conditioned on signals like text or images). While 3D generation
employs probabilistic generative models, 3D reconstruction is in most cases an ill-posed problem that requires incorporating prior knowledge and could therefore benefit from probabilistic inference methods.

In this work, we propose a probabilistic framework for 3D reconstruction. By combining a generative \textbf{prior} over latent 3D representations with a \textbf{likelihood} term from a reconstruction model, our approach predicts the full \textbf{posterior} distribution of a scene's 3D structure given sparse or noisy observations. This framework leverages a volumetric renderer based on a shared conditional neural field (CNF), trained in two stages:
\begin{enumerate}
    \item \textbf{Auto-decoding} optimizes the shared CNF and latent representations for scenes in the training set.
    \item \textbf{A diffusion model} captures the prior distribution over the latent representations.
\end{enumerate}

Our method uses a tri-plane latent structure~\citep{chanEfficientGeometryaware3D2021, Chen2022ECCV} for efficient representation, balancing global and local 3D information. Diffusion-based posterior sampling, guided by reconstruction gradients, enables probabilistic reasoning and uncertainty quantification. 
In contrast to previous work suggesting amortizing posterior
inference ~\citep{kosiorek2021nerfvae} in a variational autencoder setting, which experimentally demonstrated only limited results. By guiding the Langevin sampling with the gradient of the reconstruction
model we combine the strength of two recently successful
methods (1) iterative sampling with diffusion models and (2)
gradient based optimization for translating observations to 3D
representations.

We validate our method on tasks like single-view reconstruction, reconstruction from sparse pixels or depth, and noisy observations, demonstrating improved coverage of ground truth structures and generating uncertainty maps for unobserved regions.

Our contributions are as follows: \begin{enumerate} \item A probabilistic framework for 3D reconstruction, leveraging a diffusion prior and NeRF-based decoder.
\item A two-stage training approach: auto-decoding latent 3D representations and training a diffusion model as a prior.
\item Demonstrations on diverse 3D reconstruction tasks, showcasing robustness to sparse and noisy observations.
\item Enhanced reconstruction quality and uncertainty quantification through posterior sampling.
\end{enumerate} All code, models, and data will be released upon publication.
\section{Related Work}
\label{sec:related_work}

\textbf{Latent variable models over 3D scenes}
Early approaches like GQN~\citep{GQN} used variational autoencoders for probabilistic reasoning in simple 3D scenes but lacked specialized 3D geometry. Later, NeRF-based models~\citep{kosiorek2021nerfvae} integrated rendering pipelines but did not fully leverage NeRF's capacity for complex scenes. Our work advances this by replacing amortized inference with high-capacity diffusion models and Langevin posterior sampling. Other efforts~\citep{shen2022conditionalflownerfaccurate3d, sünderhauf2022densityawarenerfensemblesquantifying, goli2023bayesraysuncertaintyquantification} modeled uncertainty but lacked data-driven priors like ours.

\textbf{Generating 3D with 2D Generative models}
Given limited 3D ground truth data, several works~\citep{poole2022dreamfusion, watson2022novelviewsynthesisdiffusion, liu2023zero1to3} use pretrained 2D diffusion models to infer 3D representations. In a similar approach, \cite{liu2024deceptivenerf3dgsdiffusiongeneratedpseudoobservationshighquality} use a 2D prior to compute 3D uncertainty maps. While these approaches achieve impressive generative visual results, they do not explicitly reason about the 3D structure of the scene, which leads to less consistency in generation (see experiment in Sec.~\ref{sec:appendix_consistency} in the appendix) and prevents them from performing the full range of 3D probabilistic reasoning tasks, e.g. reconstruction from depth information.  

\textbf{Generative models of observed 3D representations}
Despite data scarcity, some works~\citep{shue20223dneuralfieldgeneration, erkoç2023hyperdiffusiongeneratingimplicitneural} train diffusion models directly on 3D datasets using representations like tri-planes or neural fields. Our approach uniquely avoids reliance on 3D ground truth, using only 2D datasets to generate 3D scenes.

\textbf{Generative models of latent 3D representations}
Inspired by~\citep{DBLP:journals/corr/abs-2201-12204}, we enhance conditional neural fields (CNFs) with a compressed tri-plane representation and diffusion-based posterior sampling. Similar models~\citep{bautista2022gaudi, yang2023learningdiffusionpriornerfs} focus on generative tasks rather than reconstruction, while~\citep{chen2023singlestagediffusionnerfunified} unify training stages but lack our latent compression. Concurrent work~\citep{le2024robustinversegraphicsprobabilistic} explores full posterior inference but with different objectives.

The concurrent work of \cite{le2024robustinversegraphicsprobabilistic} shares a similar motivation to ours. It focuses on specific type of noisy observations using a 3D modeling of the corruption field, and  extensively demonstrate the advantages of the full posterior distribution over the maximum only (MAP inference).
In~\cite{zhang2024gaussiancubestructuredexplicitradiance} a 3D generative model is trained based on a Gaussian splatting representation which could also be combined with posterior sampling in future research.



\section{Background}
\label{sec:background}
\subsection{Auto-decoding 3D representations}
Recent advances in 3D scene representation have leveraged deep neural networks. NeRF~\citep{mildenhall2020nerf} introduced a method to reconstruct 3D scenes by training a neural network on multi-view images, requiring separate models for each scene. Subsequent work, such as PixelNeRF~\citep{yu2021pixelnerf} and IBRNet~\citep{wang2021ibrnet}, developed generalizable models that integrate prior knowledge of 3D scenes, reducing the number of views needed. These models often rely on conditional neural fields (CNFs), where a shared neural field is conditioned on scene-specific representations.

Recent studies~\citep{DBLP:journals/corr/abs-2201-12204,bautista2022gaudi,chen2023singlestagediffusionnerfunified,yang2023learningdiffusionpriornerfs} proposed the use of CNFs to train representations of scenes that can later be used in downstream tasks.
Such models use an \emph{auto-decoding} approach~\citep{bojanowskiOptimizingLatentSpace2019,park2019deepsdflearningcontinuoussigned}, where the representations are optimized for each scene concurrently with the training of the shared CNF.

Tri-plane representations~\citep{chanEfficientGeometryaware3D2021, Chen2022ECCV} have proven effective, maintaining spatial structure and balancing global and local information. These representations condition the CNF by interpolating queried 3D positions across orthogonal planes, as shown in Fig.~\ref{fig:reconstruction_model}.

\subsection{Posterior sampling with diffusion models}
\label{diffusion}

Diffusion models, such as Denoising Diffusion Probabilistic Models (DDPM)\citep{ho2020denoising}, generate high-quality samples by reversing a forward diffusion process that progressively adds noise. Many different variants \citep{sohl2015deep, song2020score} use U-Net architectures~\citep{ronneberger2015u} to predict noise and iteratively refine samples. Given a noisy input $x_t$, the denoised sample $x_{t-1}$ and the clean estimate $\hat{x}_0$ are computed as: 
\begin{equation}
\begin{aligned}
    x_{t-1} &\sim \mathcal{N} \left( \frac{1}{\sqrt{\alpha_t}} \left(x_t-\frac{1-\alpha_t }{\sqrt{1-\bar{\alpha}_t}} \epsilon_\theta(x_t,t)\right), \; \tilde{\beta}_t I \right), \\
    \hat{x}_0(x_t, t) &= \frac{1}{\sqrt{\bar{\alpha}_t}} \left(x_t-\sqrt{1-\bar{\alpha}_t} \epsilon_\theta(x_t,t) \right).
\end{aligned}
\label{eq:denoise}
\end{equation}

Where, $\bar{\alpha}_t = \prod_{s=0}^t \alpha_s$ and $\tilde{\beta}_t$ is the noise variance.
Diffusion models are widely used as priors for image restoration tasks (e.g., denoising, inpainting)~\citep{choi2021ilvr, chung2023diffusion, kawar2022denoising}, with posterior sampling defined as:

\begin{equation} \label{eq:posterior_score}
    \nabla_{x_t} \log p_t(x_t | y) = \nabla_{x_t} \log p_t(x_t) + \nabla_{x_t} \log p_t(y | x_t)
\end{equation}

This approach combines the prior score $\nabla_{x_t} \log p_t(x_t)$ with the likelihood gradient $\nabla_{x_t} \log p_t(y | x_t)$, often referred to as guidance. While exact likelihoods typically depend on clean images $x_0$, approximations have been proposed, as detailed in Sec.~\ref{sec:posterior}.


\section{Method}
\label{method}
In this section we describe our method both at training time and at inference time. Training is based on two stages: (1) training the reconstruction model (RM) while optimizing the latent representation of the training scenes (auto-decoding), and (2) training a diffusion model over the latents as a prior. At inference time we use the trained prior and the reconstruction model to perform posterior sampling of the latents. For all implementation details please refer to Sec.~\ref{sec:appendix_imp_details} in the appendix. 

\subsection{Training the representation and reconstruction}
\label{sec:reprec}
\begin{figure}
  \centering
  \includegraphics[width=0.8\textwidth]{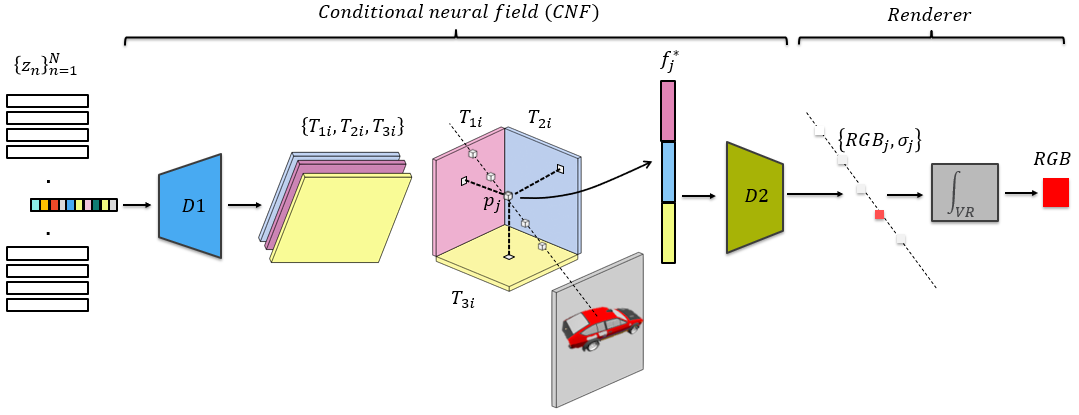}
  \caption{ The reconstruction model mapping latent representations to images of a 3D scene. The latent decoder $D1$ maps the latent vector $z_i$ corresponding to scene $i$ into three multichannel planes (tri-planes) $\{T_{1i},T_{2i},T_{3i}\}$. Given an image and a camera position from which the image was taken, a ray is projected onto the scene from each pixel of the image, and multiple 3D points are sampled along the ray. Each 3D point $p_j$, is projected onto the multichannel tri-planes where each plane produces a feature vector $f_j$ using bilinear interpolation. The three feature vectors are concatenated to form one feature vector $f_j^*$, and the decoder $D2$ is used to produce RGB and $\sigma$ values for each 3D point along the ray. Volumetric rendering is then used to generate a single RGB value to be compared to the ground truth value of the pixel in the image.} 
\label{fig:reconstruction_model}
\end{figure}

The reconstruction model is a CNF followed by a volumetric renderer. Conditioned on a scene representation, the CNF predicts the values of 3D positions within the scene that are subsequently used by the volumetric renderer. The CNF is trained while concurrently auto-decoding the representation of each scene. The role of the reconstruction model is to form a mapping from the representation vectors to the values of the observations, i.e. image pixels,  and also serve as the model through which the representation is optimized, effectively mapping the 3D scene observations back into the representation.

The model is depicted in Fig.~\ref{fig:reconstruction_model}. 
The latent vector $z_i \in \mathbb{R} ^ d$, corresponding to the $i$-th scene, is first reshaped into a 2D map of shape $r \times r \times c$.  
The latent decoder $D1$ decodes $z_i$ into a 3D tensor $T \in \mathbb{R} ^ {R \times R \times 3C}$ using a series of ResNet blocks. $T$ is reshaped to form a tri-plane representation ${T_{1i}, T_{2i}, T_{3i}} \in \mathbb{R} ^ {R \times R \times C}$.
The tri-planes structure is used for reconstruction as follows: given an image of a scene, rays are projected from each pixel into the 3D scene, and multiple 3D points are sampled along each ray. Each 3D point is projected onto the tri-planes, and using bi-linear interpolation, each plane produces a single corresponding feature vector $f \in \mathbb{R} ^ C$. The three feature vectors are concatenated to form $f^*$. The decoder $D2$, an MLP, transforms $f^*$ into $RGB$ and $\sigma$ values for the corresponding 3D position. This process is repeated for all 3D points along the ray and volumetric rendering is applied on the ray's points to generate a single RGB value for the pixel from which the ray was projected into the scene. 

The reconstruction model (RM) and the latents are trained using the auto-decoding approach as following: at each training iteration, a minibatch of scenes $\mathcal{B}$ is randomly selected along with the corresponding latent vectors, where for each scene a random set of images, and random set of pixels within the images are used. The minibatch is used to apply a forward pass of the reconstruction model on the latents, and backpropagate the loss between the model's output and ground-truth pixel values to all network weights and latent values.  
\begin{equation}
    \mathcal{L}_{rec} = \sum_{i \in \mathcal{B}} \sum_{x \in \mathcal{X}_i}  \| x - RM_\phi(z_i) \|^2
\end{equation}
where $\mathcal{B}$ is a random minibatch of scenes, and $\mathcal{X}_i$ is a random set of pixels from a random set of images from each scene $i$. The network weights are updated using $\partial \mathcal{L}_{rec} / \partial \phi$, and the latents are updated using $\partial \mathcal{L}_{rec} / \partial z_i$. In this way the latent representation for each scene is optimized while the network weights converge to their final values.
For all experiments in the paper we use a latent dimension of $1024$, which forms a highly compressed representation of the scenes. For more implementation details, see Sec.~\ref{sec:appendix_imp_details} in the appendix.

Fig.~\ref{fig:reconstruction} shows examples of reconstruction for a few selected scenes using two models that were trained on the SRN Cars~\citep{sitzmann2019srns} and Objaverse-lvis chair category~\citep{objaverse}. See Sec.~\ref{sec:appendix_data} in the appendix for details about the datasets.

After training the reconstruction models, $125$ images of held-out test scenes are used to optimize the scene latents while freezing the reconstruction model's weights, and the latents are then used to reconstruct novel views of the scenes. The results show that the latent representation captures the 3D scenes with high fidelity. In Tab.~\ref{tab:reconstruction} we compare the reconstruction accuracy of our compressed representation to ~\cite{DBLP:journals/corr/abs-2201-12204}. Our results are favorable, and we argue that this is due to the spatial structure of the tri-plane representation.

\begin{figure}
  \centering
  \begin{minipage}{0.6\textwidth}
  \centering
  \includegraphics[width=0.25\textwidth]{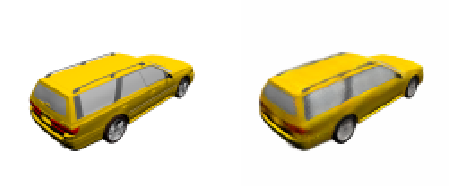}
  \includegraphics[width=0.25\textwidth]{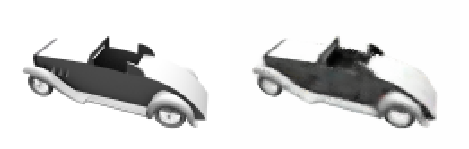}
  \includegraphics[width=0.22\textwidth]{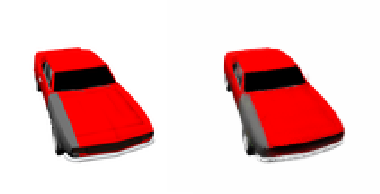}
  \includegraphics[width=0.25\textwidth]{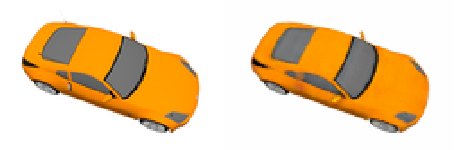}
  \\
  \includegraphics[width=0.25\textwidth]{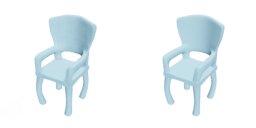}
  \includegraphics[width=0.23\textwidth]{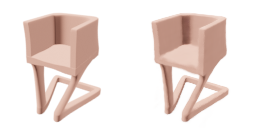}
  \includegraphics[width=0.25\textwidth]{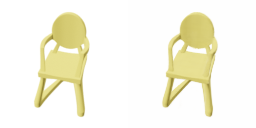}
  \includegraphics[width=0.24\textwidth]{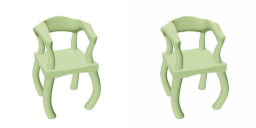}
   \end{minipage}
   \begin{minipage}{0.37\textwidth}
    \scriptsize{
\;\;\; \begin{tabular}{lcc}\toprule
\headcol & functa, \tiny{as reported in} & ours \\
\headcol & \tiny{\citep{DBLP:journals/corr/abs-2201-12204}} & \\\midrule
latent dim & 1024 & 1024 \\
PSNR train & 24.4 & 27.67 \\
PSNR test  & 23.1 & 26.9 \\\bottomrule
\end{tabular}}
\vspace{-0.2cm}
\captionof{table}{Reconstruction from latent.}
\label{tab:reconstruction}
\end{minipage}
  \caption{Novel view reconstruction for held out 3D scenes. Each pair shows the ground truth image (left) and the reconstructed image (right). Top row: SRN cars. Bottom row: Objaverse chairs.}
\label{fig:reconstruction}
\end{figure}



\subsection{Training the prior}
\label{sec:prior}

The goal of the second stage is to obtain a prior over the latent representation. This is achieved by training a generative model based on a Denoising Diffusion Probabilistic Model (DDPM)~\citep{ho2020denoising} on the latent data obtained in the first stage. 
As is standard in diffusion models, the model is based on a U-net architecture~\citep{ronneberger2015u} that is trained to denoise the latent representations $\{z_i\}_{n=1}^{N} \in \mathbb{R}^{d}$. To comply with the U-net architecture, the latents are reshaped to be $\{z_i\}_{n=1}^{N} \in \mathbb{R}^{r \times r \times c}$. The training loss is computed by:
\begin{equation}
    \mathcal{L}_{gen} = \mathbb{E}_{z \in \{z\}, \epsilon \in \mathcal{N}(0, 1), t \in U[0, T]} \;  \|\epsilon_\theta \left(\sqrt{\bar{\alpha}_t} z + \sqrt{1-\bar{\alpha}_t} \epsilon, t \right) - \epsilon \|^2
\end{equation}

\begin{figure}
  \centering
  \includegraphics[width=0.39\textwidth]{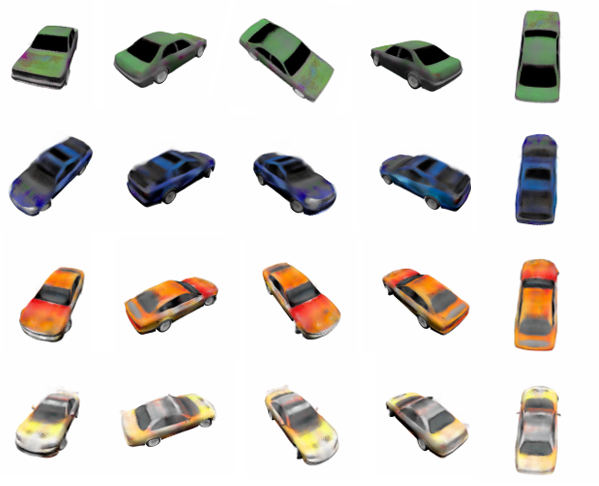} \;\;\;\;
  \includegraphics[width=0.37\textwidth]{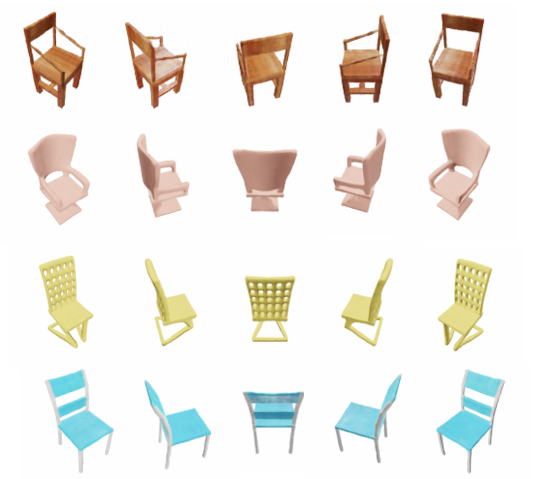}
  \caption{Samples from the trained diffusion model. Each row corresponds to a different sample of the latent representation, corresponding to a different 3D scene, and each column shows a different view reconstructed from the same scene. Left: SRN cars. Right: Objaverse-lvis chairs.}
\label{fig:diffusiongeneration}
\end{figure}

Our implementation is based on \cite{graikos2022diffusion}. Fig~\ref{fig:diffusiongeneration} shows examples of random samples generated from the learned prior. Each image is generated by first sampling a latent from the prior, corresponding to a sampled scene (rows), and then using the reconstruction model to render images of the scene from different views (columns). The resulting samples show both coherence and diversity.

\subsection{Sampling from the posterior}
\label{sec:posterior}
As described in Sec.~\ref{diffusion}, different methods have been proposed to sample from posterior distributions given a trained diffusion model as a prior. These methods consist of adding a likelihood term to each step in the iterative process of sampling from the prior. Here, the likelihood term comes from applying the reconstruction model (RM) on the estimated latent, and computing a squared loss compared to the given observation $y$ , which corresponds to a Gaussian log-likelihood. 
\begin{equation} \label{eq:rec_error}
    \log p(y | z) = - s \| y - RM_\phi(z) \|^2 + const. = - \mathcal{L}_{rec} + const.
\end{equation}
where $s$ is a scaling factor corresponding to the assumed variance of the reconstruction.   

\begin{figure}[ht]
  \centering    
  \begin{minipage}{0.42\textwidth}
  \small{
\begin{algorithm}[H]
\small{
    \caption{Posterior Sampling}
    \label{alg:posterior}
   \KwIn{images $y$, scale $s$}
   Initialize $z_T \sim N(0,1)$ \\
   \For{$t=T$ \KwTo $1$}{
        $\epsilon \gets \text{U-net}_\theta(z_t ,t)$\\
        $z_{t-1} \sim \mathcal{N}\big(\frac{1}{\sqrt{\alpha_t}} \left(z_t-\frac{1-\alpha_t }{\sqrt{1-\bar{\alpha}_t}} \epsilon\right), \tilde{\beta} \big)$\\[0.2cm]
        $\tilde{z}_0(z_{t-1}) = \frac{1}{\sqrt{\bar{\alpha}_t}} \left(z_{t-1}-\sqrt{1-\bar{\alpha}_t} \epsilon \right)$\\[0.2cm]
        $\mathcal{L}_{rec} = \left\| y - RM_\phi\left(\tilde{z}_0(z_{t-1})\right) \right\|^2 $\\[0.2cm]
        $z_{t-1} \gets z_{t-1} - s \partial \mathcal{L}_{rec} / \partial z_{t-1}$\\
   }}
\end{algorithm}}
  \end{minipage}
  \begin{minipage}{0.55\textwidth}
  \includegraphics[width=0.9\textwidth]{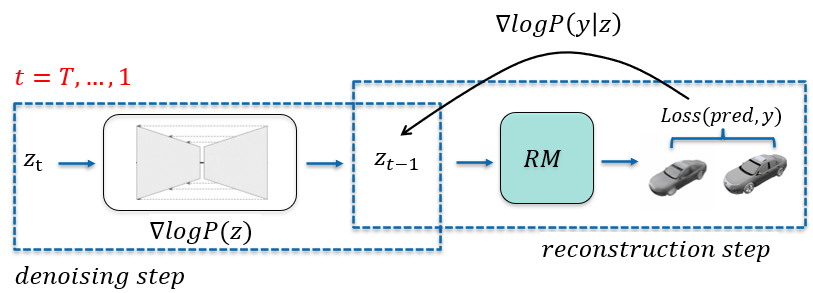}  
  \end{minipage}
  \caption{\textbf{Left:} The posterior sampling algorithm. \textbf{Right: }Illustration of a single step in the iterative process. Conditioned on the previous estimate $z_t$, the U-net predicts the noise, which is used to compute both $z_{t-1}$ and $\tilde{z}_0$. The latter is fed to the reconstruction model to predict an image from the given view which is compared to the ground truth image $y$. The error is backpropagated through the frozen networks to compute a gradient which is then added to $z_{t-1}$.}
\label{fig:posterior_sampling}
\end{figure}

The method is depicted in Fig.~\ref{fig:posterior_sampling}, and described in Alg.~\ref{alg:posterior}.
In more detail, at each step $t$ the output of the U-net $\epsilon_\theta(z_t ,t)$ is used to compute the one-step denoised latent $z_{t-1}$ and a fully denoised estimate $\hat{z_0}$ (Eq.~\ref{eq:denoise}).  The clean estimate is fed to the reconstruction model (RM) which outputs a prediction of the input views. A gradient of the log-likelihood with repsect to $z_t$ can be computed by back-propagating the reconstruction error (Eq.~\ref{eq:rec_error}) between the predicted images and the observed ground-truth images. However, this requires back-propagating through the U-net at each step. In order to accelerate inference, we approximate this gradient by computing $\tilde{z}_0(z_{t-1}) = \frac{1}{\sqrt{\bar{\alpha}_t}} \left(z_{t-1}-\sqrt{1-\bar{\alpha}_t} \epsilon_\theta(z_t ,t) \right)$,  and the gradient with respect to $z_{t-1}$. When using many sampling steps we empirically observe that the difference between $z_t$ and $z_{t-1}$ is negligible and this approximation can be used to efficiently compute the posterior score:
\begin{equation}\label{eq:app_posterior_score}
\nabla_{z_t} \log p_t(z_t\mid y)  \approx \nabla_{z_t} \log p_t(z_t) +  \nabla_{z_{t-1}} \log p\big(y\mid \tilde{z}_0 (z_{t-1}, t) \big),
\end{equation}
Repeating this process for $t = T ... 1$ forms an approximated Langevin sampling process from the posterior distribution. 

As the reconstruction loss is calculated with no regards to pixel order or quantity, this approach allows training a single prior model, and then use it to generate posterior samples for various types of conditioning signals. Examples include conditioning on many images, few images, or even a few random pixels per scene. 
Moreover, the desired inference task does not even need to be known at training time, as long as a corresponding reconstruction term can be formulated and differentiated at inference time.


\section{Experiments}
\label{sec:experiments}

\begin{figure}
\centering
\begin{tabular}{ccc}
more information & \multicolumn{2}{c}{less information\;\;\;\;\;\;\;} \\
\includegraphics[width=0.3\textwidth]{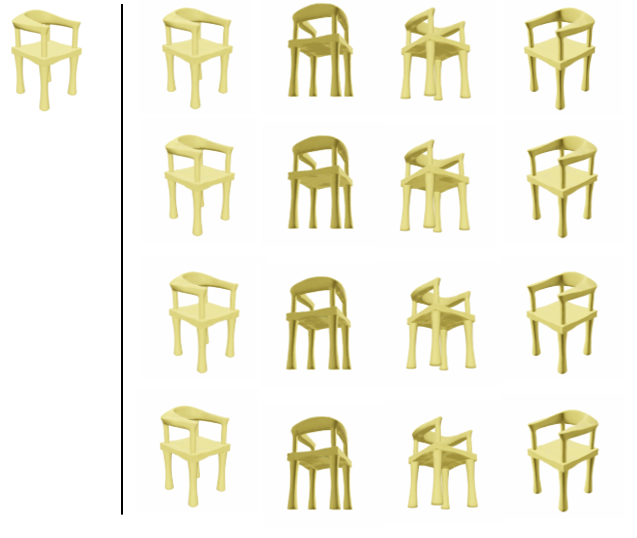} &
\includegraphics[width=0.32\textwidth]{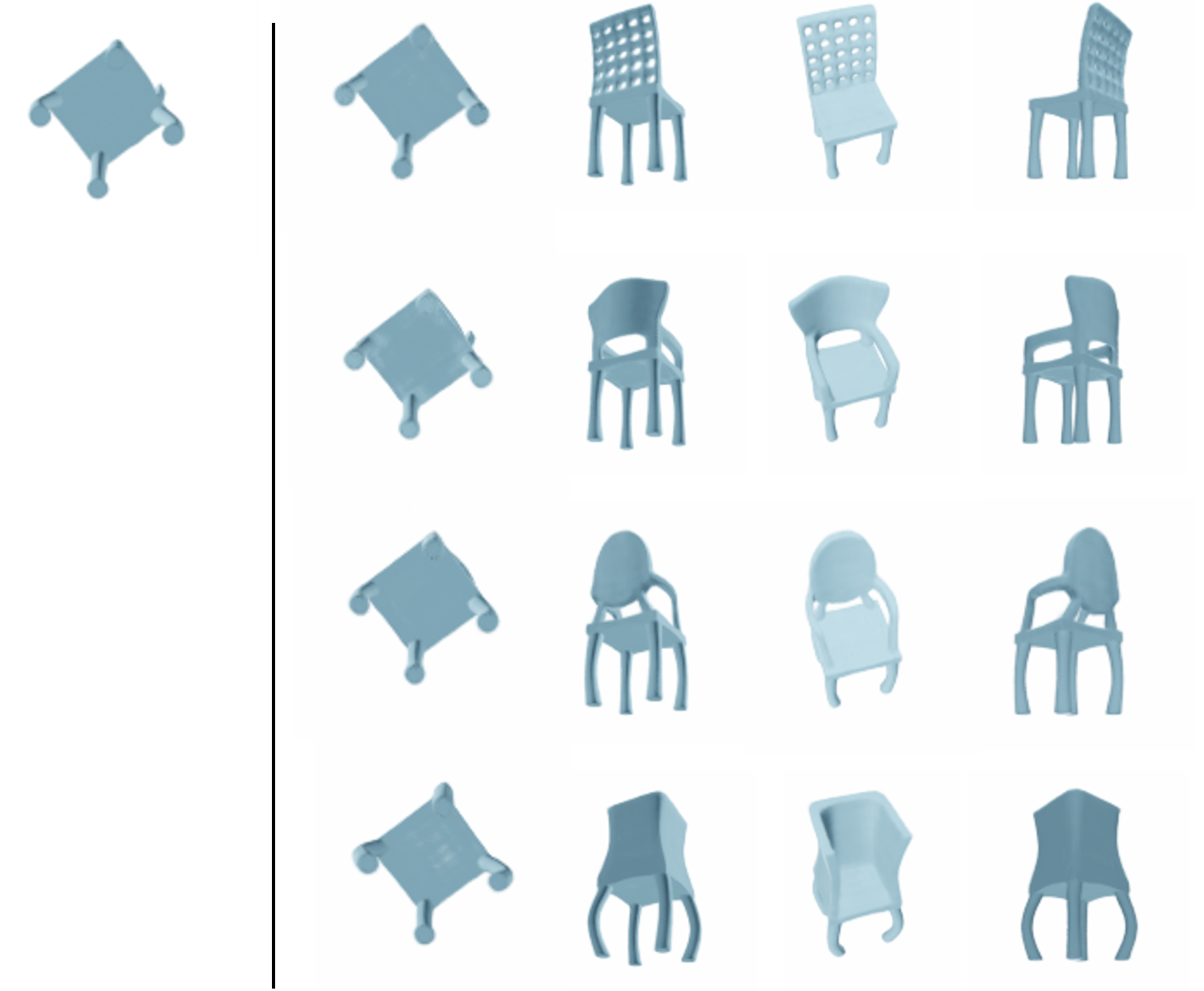} &
\includegraphics[width=0.38\textwidth]{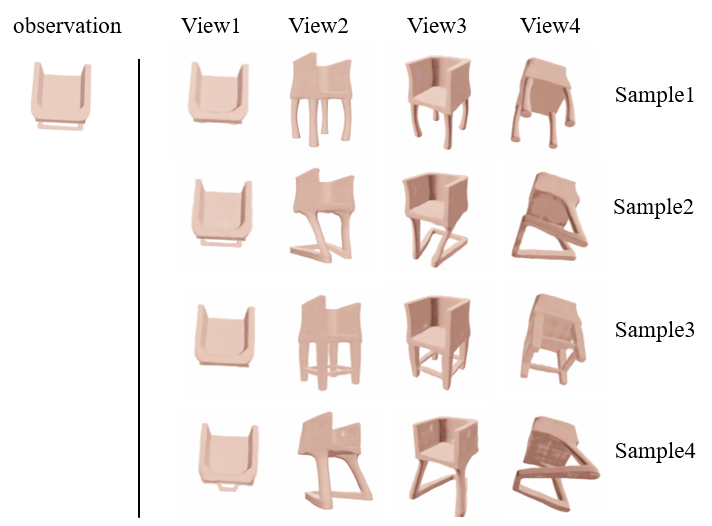} \\
\end{tabular}
    
\caption{Posterior samples given a single view for Objaverse chairs. Each row corresponds to a different sample of the scene, and each column shows a different view. The observation in the example on the left carries high information about the scene, resulting in very similar samples. The observations in the middle and right scenes are less informative, and therefore result in more diverse samples, where the chairs are completed with different possible configurations of legs, armrests and backrests. These example demonstrate a coherent merging of observed data and prior information.}
\label{fig:posteriorsampling}
\end{figure}

\begin{figure}[ht]
\centering
\scriptsize{
\begin{tabular}{l}
~\\
observed \\
images \\[0.4cm]
\\
TensoRF \\
\tiny{\citep{Chen2022ECCV}} \\
(80 images)\\[0.3cm]
\\
Posterior \\
Sampling \\
(5 images)
\end{tabular}}
\begin{tabular}{cc}
$\sigma=0.04 \;\;\; \sigma=0.2 \;\;\;\; \sigma=0.4 \;\;\; \sigma=0.8$ & $\sigma=0.04 \;\;\; \sigma=0.2 \;\;\;\; \sigma=0.4 \;\;\; \sigma=0.8$ \\
\includegraphics[width=0.35\textwidth]{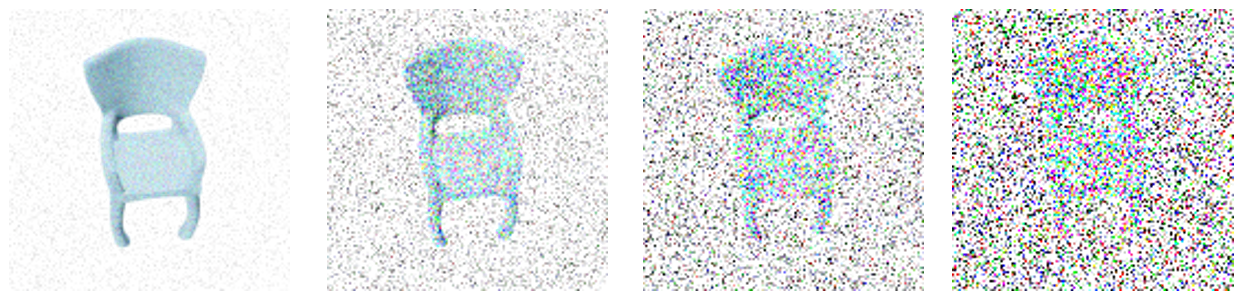} &
  \includegraphics[width=0.35\textwidth]{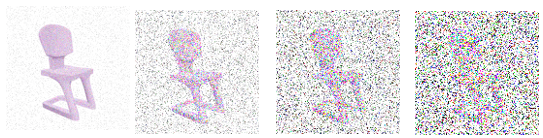}
  \\
  \includegraphics[width=0.35\textwidth]{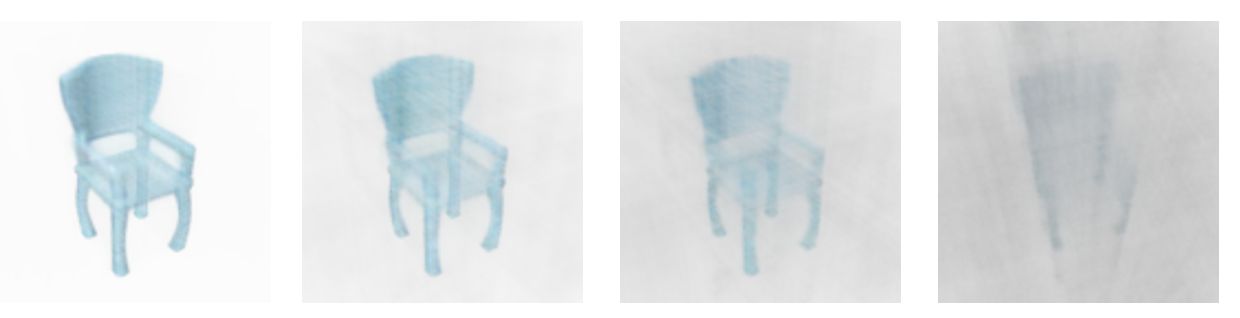} &
  \includegraphics[width=0.35\textwidth]{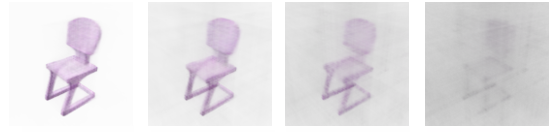}
  \\
  \includegraphics[width=0.35\textwidth]{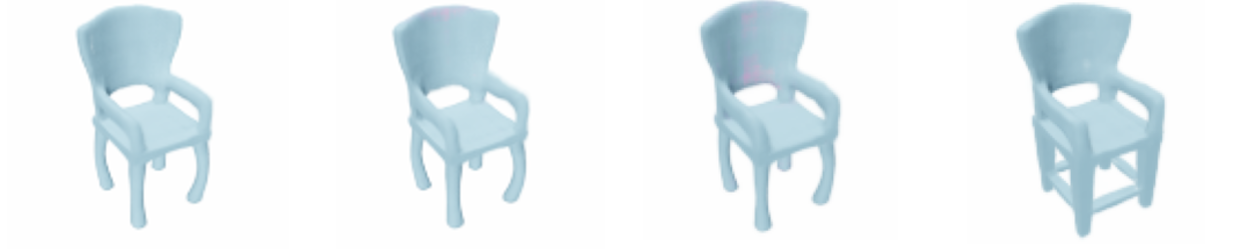} &
  \includegraphics[width=0.35\textwidth]{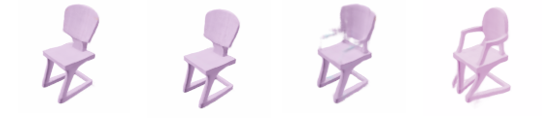}
  \\
  \end{tabular}
  \caption{3D reconstruction from noisy images. Reconstruction from 80 images without a prior (TensoRF) quickly deteriorates as noise increases. Using our prior to perform posterior sampling results in a much more robust method, significantly outperforming TensoRF even when using an order of magnitude less images (5 images).}
  \label{fig:noise_vanilla_vs_guidance}
\end{figure}

For all experiments we use the same model and the same configuration.

\subsubsection*{Generating conditional samples}
We show results of generating posterior samples given one observed image per scene. In Fig.~\ref{fig:posteriorsampling}, three examples from Objaverse chairs are shown. In the scene shown on the left, the given image contains enough information to predict any view of the scene with certainty. This results in multiple samples (rows) that are almost identical.  In the other examples the observed image is less informative and does not provide enough information about the scene from all angles. Therefore, samples from the posterior exhibit more diversity in the way they complete the missing information.  More concretely, the chairs observed from uninformative views are predicted to have different possible leg, armrest and backrest configurations. Note that while the samples are different, the generated latent is a 3D representation, so each sample can be used to predict a coherent set of images from different views. 

In Fig.~\ref{fig:partialguidance}, we demonstrate the ability of the method to perform more diverse probabilistic reasoning tasks. We show prediction from half-image inputs, from a sparse set of pixels of one image (5\%), and from a sparse set of depth map pixels (5\% of a depth map from a single view). For each scene we show three samples, showing two different views for each, and an uncertainty map. 
The uncertainty is computed by generating 10 samples of the scene, rendering corresponding 10 images for each view and computing the variance in the rendered images.
Using our method for partial RGB observations (half-image or sparse pixels) is trivial to implement since the reconstruction model operates per pixel and can be used to predict any subset of pixels in the scene. 
In the case of depth data, we implement a different reconstruction loss comparing the predicted $\sigma$ values to ground truth values without using the $RGB$ prediction and the renderer in Fig. \ref{fig:reconstruction_model}. Given a depth pixel value, the ground truth value of $\sigma$ is set to $1$ for the 3D point on the ray sampled at the given depth value, and $0$ for all the other 3D points.
We emphasize that this reconstruction model is formulated at inference time and is not used at training.
The results show the different plausible predictions of the scene and the resulting uncertainty. For the first case we see that the uncertainty is high for the hidden half of the scene as expected. 
For the other two cases, samples generated from sparse pixel observations demonstrate a high degree of similarity, suggesting, perhaps surprisingly, that even just 5\% of the pixels from a single view is sufficient for accurate 3D scene prediction. In case of the sparse depth data, the only uncertain aspect is the object color. 

In Fig.~\ref{fig:noise_vanilla_vs_guidance}, we evaluate the robustness of our method in 3D reconstruction from noisy images by comparing it to TensoRF~\citep{Chen2022ECCV} without a prior. By generating samples from multiple scene images under increasing noise levels, we demonstrate that posterior sampling significantly improves resilience to noise. While TensoRF, trained on 80 images, experiences a sharp performance drop as noise increases, our method maintains stable performance even with only 5 input images.


\section{Conclusion}
\label{sec:conclusion}

In this work we introduced a methodology that combines the strengths of NeRF-based 3D reconstruction together with the probabilistic reasoning of diffusion models. 
Our method views 3D reconstruction as an ill-posed perception problem that requires reconciling the observed information with prior knowledge.
We showed that (1) 3D scenes can be efficiently represented by compact latent vectors, using a reconstruction model that consists of a tri-plane representation, which preserves spatial structure within the 3D model; and (2) this representation is amenable to training a strong diffusion-model based prior that can later be used to solve various inference tasks. 
We highlight the importance of predicting the full posterior distribution rather than optimizing for an average sample with higher PSNR (see Sec.~\ref{sec:appendix_psnr}). Averaging tends to produce oversmoothed results that may score well numerically but fail to capture the full variability of plausible 3D structures. By emphasizing diverse posterior samples, our approach better represents the inherent uncertainty in 3D scene synthesis, leading to more robust and generalizable models and solving various 3D reconstruction tasks.


\textbf{Limitations and future work}: 
A main challenge that remains in 3D reconstruction is scaling to more complex and more diverse data towards developing methods that can reliably predict real 3D scenes from different levels of observations. Another challenge is the slow sampling time with diffusion models. 
While our results are demonstrated on small scale data, we believe that the compressed representation  and the principled way of handling uncertainty that we propose, combined with recent developments in accelerating diffusion model sampling, are key for scaling up these models to larger and more complex datasets.

\section*{Acknowledgments}
We gratefully acknowledge the PhD funding provided by the data science research center (DSRC) at the university of Haifa, Israel.

\bibliography{iclr2025_conference}

@misc{yu2021pixelnerf,
      title={pixelNeRF: Neural Radiance Fields from One or Few Images}, 
      author={Alex Yu and Vickie Ye and Matthew Tancik and Angjoo Kanazawa},
      year={2021},
      eprint={2012.02190},
      archivePrefix={arXiv},
      primaryClass={cs.CV}
}

@inproceedings{mildenhall2020nerf,
 title={NeRF: Representing Scenes as Neural Radiance Fields for View Synthesis},
 author={Ben Mildenhall and Pratul P. Srinivasan and Matthew Tancik and Jonathan T. Barron and Ravi Ramamoorthi and Ren Ng},
 year={2020},
 booktitle={ECCV},
}

@INPROCEEDINGS{Chen2022ECCV,
              author = {Anpei Chen and Zexiang Xu and Andreas Geiger and Jingyi Yu and Hao Su},
              title = {TensoRF: Tensorial Radiance Fields},
              booktitle = {European Conference on Computer Vision (ECCV)},
              year = {2022}
            }

@inproceedings{graikos2022diffusion,
   title={Diffusion Models as Plug-and-Play Priors},
   author={Alexandros Graikos and Nikolay Malkin and Nebojsa Jojic and Dimitris Samaras},
   booktitle={Thirty-Sixth Conference on Neural Information Processing Systems},
   year={2022},
   url={https://arxiv.org/pdf/2206.09012.pdf}
}

@misc{bautista2022gaudi,
      title={GAUDI: A Neural Architect for Immersive 3D Scene Generation}, 
      author={Miguel Angel Bautista and Pengsheng Guo and Samira Abnar and Walter Talbott and Alexander Toshev and Zhuoyuan Chen and Laurent Dinh and Shuangfei Zhai and Hanlin Goh and Daniel Ulbricht and Afshin Dehghan and Josh Susskind},
      year={2022},
      eprint={2207.13751},
      archivePrefix={arXiv},
      primaryClass={cs.CV}
}

@misc{DBLP:journals/corr/abs-2201-12204,
      title={From data to functa: Your data point is a function and you can treat it like one}, 
      author={Emilien Dupont and Hyunjik Kim and S. M. Ali Eslami and Danilo Rezende and Dan Rosenbaum},
      year={2022},
      eprint={2201.12204},
      archivePrefix={arXiv},
      primaryClass={cs.LG},
      url={https://arxiv.org/abs/2201.12204}, 
}

@misc{kosiorek2021nerfvae,
      title={NeRF-VAE: A Geometry Aware 3D Scene Generative Model}, 
      author={Adam R. Kosiorek and Heiko Strathmann and Daniel Zoran and Pol Moreno and Rosalia Schneider and Soňa Mokrá and Danilo J. Rezende},
      year={2021},
      eprint={2104.00587},
      archivePrefix={arXiv},
      primaryClass={stat.ML}
}

@misc{ho2020denoising,
      title={Denoising Diffusion Probabilistic Models}, 
      author={Jonathan Ho and Ajay Jain and Pieter Abbeel},
      year={2020},
      eprint={2006.11239},
      archivePrefix={arXiv},
      primaryClass={cs.LG}
}

@inproceedings{sitzmann2019srns,
        author = {Sitzmann, Vincent
                    and Zollh{\"o}fer, Michael
                    and Wetzstein, Gordon},
        title = {Scene Representation Networks: Continuous 3D-Structure-Aware Neural Scene Representations},
	booktitle = {Advances in Neural Information Processing Systems},
        year={2019}
    }

@misc{poole2022dreamfusion,
      title={DreamFusion: Text-to-3D using 2D Diffusion}, 
      author={Ben Poole and Ajay Jain and Jonathan T. Barron and Ben Mildenhall},
      year={2022},
      eprint={2209.14988},
      archivePrefix={arXiv},
      primaryClass={cs.CV}
}

@misc{liu2023zero1to3,
      title={Zero-1-to-3: Zero-shot One Image to 3D Object}, 
      author={Ruoshi Liu and Rundi Wu and Basile Van Hoorick and Pavel Tokmakov and Sergey Zakharov and Carl Vondrick},
      year={2023},
      eprint={2303.11328},
      archivePrefix={arXiv},
      primaryClass={cs.CV}
}

@misc{wang2021ibrnet,
      title={IBRNet: Learning Multi-View Image-Based Rendering}, 
      author={Qianqian Wang and Zhicheng Wang and Kyle Genova and Pratul Srinivasan and Howard Zhou and Jonathan T. Barron and Ricardo Martin-Brualla and Noah Snavely and Thomas Funkhouser},
      year={2021},
      eprint={2102.13090},
      archivePrefix={arXiv},
      primaryClass={cs.CV}
}

@misc{goli2023bayesraysuncertaintyquantification,
      title={Bayes' Rays: Uncertainty Quantification for Neural Radiance Fields}, 
      author={Lily Goli and Cody Reading and Silvia Sellán and Alec Jacobson and Andrea Tagliasacchi},
      year={2023},
      eprint={2309.03185},
      archivePrefix={arXiv},
      primaryClass={cs.CV},
      url={https://arxiv.org/abs/2309.03185}, 
}

@misc{shen2022conditionalflownerfaccurate3d,
      title={Conditional-Flow NeRF: Accurate 3D Modelling with Reliable Uncertainty Quantification}, 
      author={Jianxiong Shen and Antonio Agudo and Francesc Moreno-Noguer and Adria Ruiz},
      year={2022},
      eprint={2203.10192},
      archivePrefix={arXiv},
      primaryClass={cs.CV},
      url={https://arxiv.org/abs/2203.10192}, 
}

@misc{sünderhauf2022densityawarenerfensemblesquantifying,
      title={Density-aware NeRF Ensembles: Quantifying Predictive Uncertainty in Neural Radiance Fields}, 
      author={Niko Sünderhauf and Jad Abou-Chakra and Dimity Miller},
      year={2022},
      eprint={2209.08718},
      archivePrefix={arXiv},
      primaryClass={cs.CV},
      url={https://arxiv.org/abs/2209.08718}, 
}

@misc{liu2024deceptivenerf3dgsdiffusiongeneratedpseudoobservationshighquality,
      title={Deceptive-NeRF/3DGS: Diffusion-Generated Pseudo-Observations for High-Quality Sparse-View Reconstruction}, 
      author={Xinhang Liu and Jiaben Chen and Shiu-hong Kao and Yu-Wing Tai and Chi-Keung Tang},
      year={2024},
      eprint={2305.15171},
      archivePrefix={arXiv},
      primaryClass={cs.CV},
      url={https://arxiv.org/abs/2305.15171}, 
}

@misc{watson2022novelviewsynthesisdiffusion,
      title={Novel View Synthesis with Diffusion Models}, 
      author={Daniel Watson and William Chan and Ricardo Martin-Brualla and Jonathan Ho and Andrea Tagliasacchi and Mohammad Norouzi},
      year={2022},
      eprint={2210.04628},
      archivePrefix={arXiv},
      primaryClass={cs.CV},
      url={https://arxiv.org/abs/2210.04628}, 
}

@misc{zhang2024gaussiancubestructuredexplicitradiance,
      title={GaussianCube: A Structured and Explicit Radiance Representation for 3D Generative Modeling}, 
      author={Bowen Zhang and Yiji Cheng and Jiaolong Yang and Chunyu Wang and Feng Zhao and Yansong Tang and Dong Chen and Baining Guo},
      year={2024},
      eprint={2403.19655},
      archivePrefix={arXiv},
      primaryClass={cs.CV},
      url={https://arxiv.org/abs/2403.19655}, 
}

@String(ECCV= {Eur. Conf. Comput. Vis.})

@String(NIPS= {Adv. Neural Inform. Process. Syst.})

@String(ICLR = {Int. Conf. Learn. Represent.})

@String(ECCV  = {ECCV})

@String(NIPS  = {NeurIPS})

@String(ICLR  = {ICLR})

@inproceedings{chung2023diffusion,                    
 title={Diffusion Posterior Sampling for General Noisy Inverse Problems},                    
 author={Hyungjin Chung and Jeongsol Kim and Michael Thompson Mccann and Marc Louis Klasky and Jong Chul Ye},                    
 booktitle=ICLR,                  
 year={2023},                    
 url={https://openreview.net/forum?id=OnD9zGAGT0k}                    
}

@article{choi2021ilvr,
  title={Ilvr: Conditioning method for denoising diffusion probabilistic models},
  author={Choi, Jooyoung and Kim, Sungwon and Jeong, Yonghyun and Gwon, Youngjune and Yoon, Sungroh},
  journal={arXiv preprint arXiv:2108.02938},
  year={2021}
}

@article{kawar2022denoising,
  title={Denoising diffusion restoration models},
  author={Kawar, Bahjat and Elad, Michael and Ermon, Stefano and Song, Jiaming},
  journal=NIPS,
  volume={35},
  pages={23593--23606},
  year={2022}
}

@inproceedings{ronneberger2015u,
  title={U-net: Convolutional networks for biomedical image segmentation},
  author={Ronneberger, Olaf and Fischer, Philipp and Brox, Thomas},
  booktitle={Medical Image Computing and Computer-Assisted Intervention (MICCAI)},
  pages={234--241},
  year={2015}
}

@inproceedings{sohl2015deep,
  title={Deep unsupervised learning using nonequilibrium thermodynamics},
  author={Sohl-Dickstein, Jascha and Weiss, Eric and Maheswaranathan, Niru and Ganguli, Surya},
  booktitle={Int. conf. machine learning},
  pages={2256--2265},
  year={2015}
}

@article{song2020score,
  title={Score-based generative modeling through stochastic differential equations},
  author={Song, Yang and Sohl-Dickstein, Jascha and Kingma, Diederik P and Kumar, Abhishek and Ermon, Stefano and Poole, Ben},
  journal={arXiv preprint arXiv:2011.13456},
  year={2020}
}

@misc{yang2023learningdiffusionpriornerfs,
      title={Learning a Diffusion Prior for NeRFs}, 
      author={Guandao Yang and Abhijit Kundu and Leonidas J. Guibas and Jonathan T. Barron and Ben Poole},
      year={2023},
      eprint={2304.14473},
      archivePrefix={arXiv},
      primaryClass={cs.CV},
      url={https://arxiv.org/abs/2304.14473}, 
}

@article{objaverse,
  title={Objaverse: A Universe of Annotated 3D Objects},
  author={Matt Deitke and Dustin Schwenk and Jordi Salvador and Luca Weihs and
          Oscar Michel and Eli VanderBilt and Ludwig Schmidt and
          Kiana Ehsani and Aniruddha Kembhavi and Ali Farhadi},
  journal={arXiv preprint arXiv:2212.08051},
  year={2022}
}

@misc{park2019deepsdflearningcontinuoussigned,
      title={DeepSDF: Learning Continuous Signed Distance Functions for Shape Representation}, 
      author={Jeong Joon Park and Peter Florence and Julian Straub and Richard Newcombe and Steven Lovegrove},
      year={2019},
      eprint={1901.05103},
      archivePrefix={arXiv},
      primaryClass={cs.CV},
      url={https://arxiv.org/abs/1901.05103}, 
}

@article{chanEfficientGeometryaware3D2021,
  title = {Efficient {{Geometry-aware 3D Generative Adversarial Networks}}},
  author = {Chan, Eric R and Lin, Connor Z and Chan, Matthew A and Nagano, Koki and Pan, Boxiao and De Mello, Shalini and Gallo, Orazio and Guibas, Leonidas and Tremblay, Jonathan and Khamis, Sameh and others},
  year = {2021},
  journal = {arXiv preprint arXiv:2112.07945},
  eprint = {2112.07945},
  archiveprefix = {arXiv}
}

@misc{le2024robustinversegraphicsprobabilistic,
      title={Robust Inverse Graphics via Probabilistic Inference}, 
      author={Tuan Anh Le and Pavel Sountsov and Matthew D. Hoffman and Ben Lee and Brian Patton and Rif A. Saurous},
      year={2024},
      eprint={2402.01915},
      archivePrefix={arXiv},
      primaryClass={cs.CV},
      url={https://arxiv.org/abs/2402.01915}, 
}

@misc{shue20223dneuralfieldgeneration,
      title={3D Neural Field Generation using Triplane Diffusion}, 
      author={J. Ryan Shue and Eric Ryan Chan and Ryan Po and Zachary Ankner and Jiajun Wu and Gordon Wetzstein},
      year={2022},
      eprint={2211.16677},
      archivePrefix={arXiv},
      primaryClass={cs.CV},
      url={https://arxiv.org/abs/2211.16677}, 
}

@article{GQN,
author = {Eslami, S. and Jimenez Rezende, Danilo and Besse, Frederic and Viola, Fabio and Morcos, Ari and Garnelo, Marta and Ruderman, Avraham and Rusu, Andrei and Danihelka, Ivo and Gregor, Karol and Reichert, David and Buesing, Lars and Weber, Theophane and Vinyals, Oriol and Rosenbaum, Dan and Rabinowitz, Neil and King, Helen and Hillier, Chloe and Botvinick, Matt and Hassabis, Demis},
year = {2018},
month = {06},
pages = {1204-1210},
title = {Neural scene representation and rendering},
volume = {360},
journal = {Science},
doi = {10.1126/science.aar6170}
}

@misc{erkoç2023hyperdiffusiongeneratingimplicitneural,
      title={HyperDiffusion: Generating Implicit Neural Fields with Weight-Space Diffusion}, 
      author={Ziya Erkoç and Fangchang Ma and Qi Shan and Matthias Nießner and Angela Dai},
      year={2023},
      eprint={2303.17015},
      archivePrefix={arXiv},
      primaryClass={cs.CV},
      url={https://arxiv.org/abs/2303.17015}, 
}

@misc{chen2023singlestagediffusionnerfunified,
      title={Single-Stage Diffusion NeRF: A Unified Approach to 3D Generation and Reconstruction}, 
      author={Hansheng Chen and Jiatao Gu and Anpei Chen and Wei Tian and Zhuowen Tu and Lingjie Liu and Hao Su},
      year={2023},
      eprint={2304.06714},
      archivePrefix={arXiv},
      primaryClass={cs.CV},
      url={https://arxiv.org/abs/2304.06714}, 
}

@misc{bojanowskiOptimizingLatentSpace2019,
  title = {Optimizing the {{Latent Space}} of {{Generative Networks}}},
  author = {Bojanowski, Piotr and Joulin, Armand and {Lopez-Paz}, David and Szlam, Arthur},
  year = {2019},
  month = may,
  number = {arXiv:1707.05776},
  eprint = {1707.05776},
  primaryclass = {cs, stat},
  publisher = {arXiv},
  doi = {10.48550/arXiv.1707.05776},
  url = {http://arxiv.org/abs/1707.05776},
  urldate = {2024-09-23},
  archiveprefix = {arXiv}
}
\bibliographystyle{iclr2025_conference}

\newpage
\appendix

\section{Data}
\label{sec:appendix_data}
We use two datasets in our experiments. 
The first dataset is SRN Cars~\citep{sitzmann2019srns}, which comprises 3,200 scenes with 250 images each. We randomly divide the images in each scene evenly between training images and test images, and we use 3,000 scenes for training, holding out 200 scenes for testing.
The second dataset we use is the Objaverse-lvis chair category~\citep{objaverse}, which comprises 439 instances with 100 images generated for each scene. While this dataset is smaller, it is more diverse in terms of shapes. We use 80\% of the images in each scene for training, and hold out $8$ scenes for testing and visualizations. For both datasets we use image resolution of $128 \times 128$.

\section{Implementation details}
\label{sec:appendix_imp_details}
In this section we describe the implementation details of our model and experiments. All code, models and data will be made available upon publication.

\subsubsection*{Training the Representation and Reconstruction model}   

The weights of the reconstruction model $\phi$ are randomly initialized, while the latent representations $z_i$ are initialized to zero. The size of the data set corresponds to the number of latent vectors, each latent representing a single scene $\{z_i\}_{i=1}^N$ (N scenes = N latents).

During training, the images of each scene optimize only its respective latent, while the entire model, including decoders, is jointly trained.   

The latent representation $z_i$ dimensions are $d=1024$, $r=16, c=4$. D1 is constructed using a series of six ResNet blocks where at each block the number of channels is the following: [4, 32, 64, 96, 128, 192]. Blocks are followed by a self-attention layer and alternating upsampling. The resulting 3D tensor $T$ is divided into two tensors responsible for generating RGB and density, $T_{RGB} \in \mathbb{R} ^ {R \times R \times 3C_{RGB}}$, $T_{\sigma} \in \mathbb{R} ^ {R \times R \times 3C_{\sigma}}$, respectively.  $T_{RGB}$ is reshaped to form a tri-plane representation ${T_{1i}, T_{2i}, T_{3i}} \in \mathbb{R} ^ {R \times R \times C_{RGB}}$. Similarly, $T_{\sigma}$ forms a tri-plane representation ${T'_{1i}, T'_{2i}, T'_{3i}} \in \mathbb{R} ^ {R \times R \times C_{\sigma}}$. Dimensions are $R=128, C_{RGB}=48, C_{\sigma}=16$. \\
For each scene $i$, we randomly select 4096 rays from pixels in the training images. Along each ray, we sample 220 3D points and project them onto the tri-planes of both the RGB and density planes separately. \\
For each (RGB and density), this projection extracts three feature vectors from the three planes for further processing.
Three vectors are concatenated into a single feature vector $f_{RGB}^* \in \mathbb{R} ^ {3C_{RGB}}$ for RGB and $f_{\sigma}^* \in \mathbb{R} ^ {3C_{\sigma}}$ for density. 
While the density feature vector $f^*_{\sigma}$ produces density for 3D points by simply summing its elements, the RGB feature vector $f^*_{RGB}$ is passed through D2 to produce a single RGB value. D2 is an MLP of 7 layers. Once all 3D points along the ray have RGB and density values, volumetric rendering, a parameterless process, produces a single RGB value to be compared with the pixel's color. 

We train the model with a minibatch $\mathcal{B}$ size of 2 scenes, and with an Adam optimizer using three different learning rates: 1e-3 for the latents, 1e-4 for the D1 parameters and $1e-3$ for D2 parameters.

Our model is based on the code published in \cite{Chen2022ECCV}.

At test time, a new latent (initialized to zeros) is coupled with the new scene and optimized using the learned/frozen decoders.

\subsubsection*{Training the Prior} 
As in Sec.\ref{sec:reprec}, latent representation $z_i$ dimensions are $d=1024$, $r=16, c=4$. The diffusion model used is implemented by \cite{graikos2022diffusion} with the following parameters: The noise scheduler is a linear schedule with parameters $T=1000, \beta_0 = 1e^{-4}, \beta_T = 2e^{-2}$. The U-net parameters are $model\_channels=64$, $num\_resnet\_blocks=2$, $channel\_mult=(1, 2, 3, 4)$, $attention\_resolutions=[8, 4]$, $num\_heads=4$. We train the model with a minibatch $\mathcal{B}$ size of 32 scenes, and with an Adam optimizer with learning rate equal to 1e-3.

The reconstruction model and the diffusion model were trained on an NVIDIA GeForce RTX 4090 for Approximately one day each.

\section{Experiments}

\subsection*{Generating 3D with 2D generative models}
\label{sec:appendix_consistency}

As mentioned in Sec.~\ref{sec:related_work}, 2D generative models for 3D generation approaches do not fully capture the underlying 3D structure of a scene. To evaluate the impact of explicit 3D structure reasoning, we trained a Neural Radiance Field (NeRF) on images generated by both standard 2D diffusion models and our proposed model, which incorporates an inherent understanding of 3D structure. Tab.~\ref{tab:2ddiffusioncomp} results indicate that NeRF trained on images from our model produces more consistent 3D reconstructions, highlighting the importance of explicit 3D reasoning in generative models for robust 3D scene synthesis.

\begin{table}[ht]
\centering
\scriptsize{
\begin{tabular*}{\columnwidth}{@{\extracolsep{\fill}}lcccl}
\toprule
\textbf{Method} & \textbf{PSNR}$\uparrow$ & \textbf{SSIM}$\uparrow$  \\

\midrule
~~3Dim (SRN cars)  & 28.53 & 0.96  \\
~~Ours (SRN cars) & 34.7 & 0.98  \\
\midrule
~~zero123 (Objaverse chairs)  & 26.8 & 0.925 \\
~~Ours (Objaverse chairs) & 43.4 & 0.99   \\
\bottomrule
\end{tabular*}
}
\vspace{0.1cm}
\caption{3D Consistency Comparison: To evaluate 3D consistency, we compare our model with 3Dim ~\citep{watson2022novelviewsynthesisdiffusion} and Zero-1-to-3 ~\citep{liu2023zero1to3}, trained on SRN Cars and Objaverse Chairs, respectively. Given an input image of a scene, each model generates multiple novel views, which are then used to train a TensoRF (NeRF) model. Since higher 3D consistency in the generated images facilitates NeRF training, models producing more consistent views enable NeRF to achieve a higher PSNR. Our results demonstrate that NeRF trained on images from our model attains the highest PSNR, highlighting the benefits of our model’s built-in 3D structural understanding for improved 3D scene synthesis.}
\label{tab:2ddiffusioncomp}
\end{table}

\subsection*{Posterior sampling}

Posterior sampling involves two types of computations: 1) denoising, using the diffusion as a prior to generate a plausible latent,  and 2) reconstruction, using the reconstruction model to align the latent with the observed views.

For all experiments we use the same model using the same inference process. We generate posterior samples using $1000$ iterations as described in Alg.~\ref{alg:posterior} with the same scale factor $s = $5e-3 for all experiments.
The only exception is the experiment with noisy data in Fig~\ref{fig:noise_vanilla_vs_guidance}, where the scale factor for most extreme noise level $\sigma=0.8$ was decreased to a value of $s = $3e-3, corresponding to the high noise variance in the observation. 


\subsubsection*{Full posterior vs higher psnr}
\label{sec:appendix_psnr}

\FloatBarrier
\begin{figure}[t]  
    \centering    \includegraphics[width=0.7\textwidth]{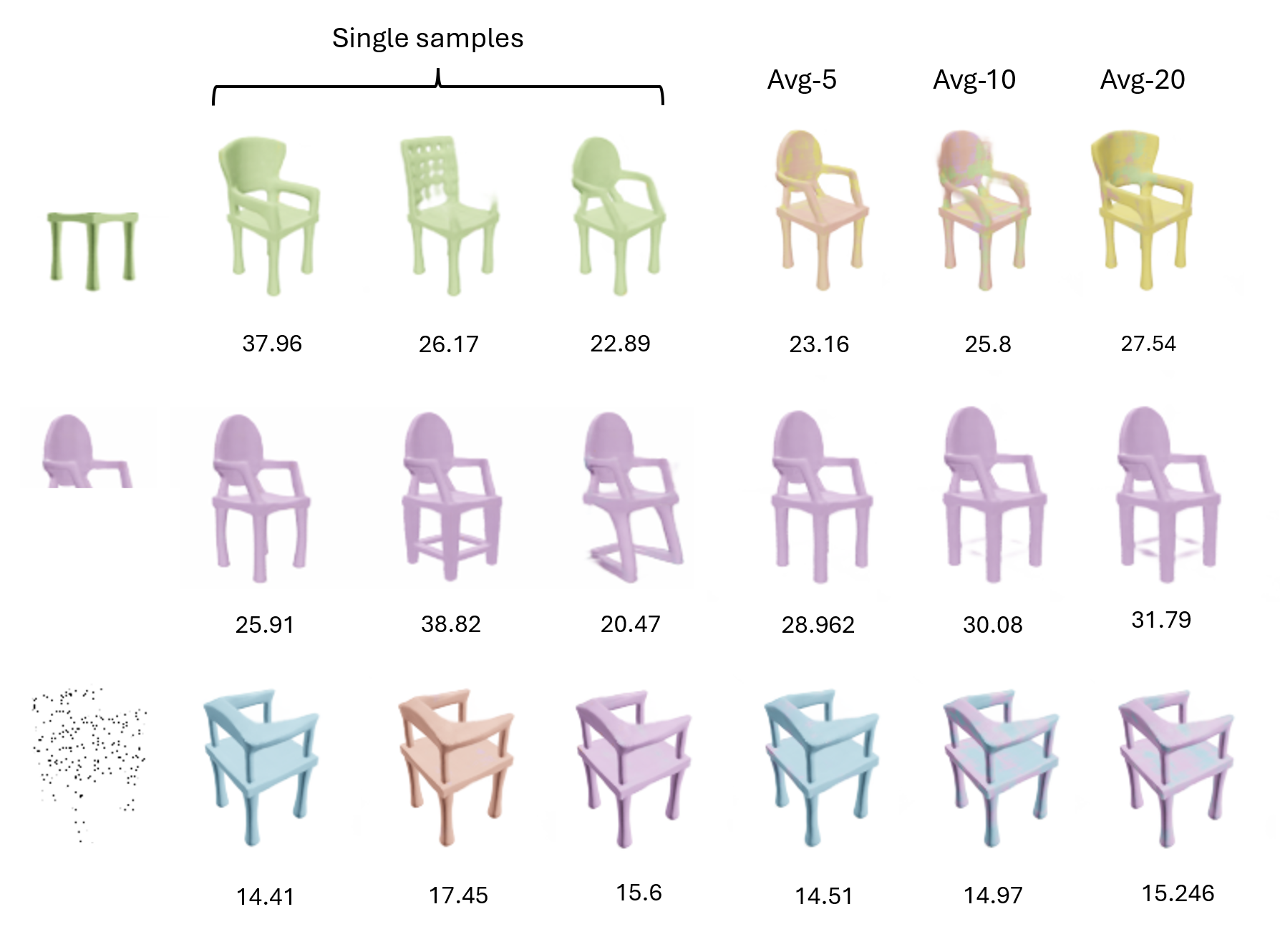}
    \caption{Averaging latent samples results in higher PSNR scores but fails to capture the full posterior distribution. In the first row, the bottom half of an image is used as guidance, while in the second row, the top half is used. Since most generated samples closely match the ground truth, averaging leads to a higher PSNR but collapses the distribution into a single reconstruction, limiting the diversity of plausible outcomes. In the third row, guidance is provided using only a few pixels from a depth image. The generated samples vary in color, as color information is not available from the depth input. Averaging across multiple samples produces an intermediate color that achieves a higher PSNR but fails to reflect the full posterior, which contains diverse color variations.}
    \label{fig:posteriorvspsnr}
\end{figure}

In Fig.~\ref{fig:posteriorvspsnr}, we present two examples (rows) of conditional posterior sampling, where the leftmost image serves as the observed image. The reconstructions are displayed alongside their corresponding PSNR values. We generate 20 samples from the posterior distribution and showcase three individual samples under "single samples," highlighting the variability in reconstruction quality—some being closer to the ground truth than others. Additionally, we display the averaged reconstructions using 5, 10, and 20 latent samples. While averaging improves numerical PSNR scores, it often leads to oversmoothed results that fail to capture the full diversity of plausible 3D structures.

\end{document}